# A Machine Learning Approach for Honey Adulteration Detection using Mineral Element Profiles

Mokhtar A. Al-Awadhi[1] and Ratnadeep R. Deshmukh[2]

[1] Department of Information Technology, Faculty of Engineering and Information Technology, Taiz University
[1,2] Department of Computer Science and IT, Dr. Babasaheb Ambedkar Marathwada University, Aurangabad, India
[1] mokhtar.awadhi@gmail.com, [2] rrdeshmukh.csit@bamu.ac.in

**Abstract.** This paper aims to develop a Machine Learning (ML)-based system for detecting honey adulteration utilizing honey mineral element profiles. The proposed system comprises two phases: preprocessing and classification. The preprocessing phase involves the treatment of missing-value attributes and normalization. In the classification phase, we use three supervised ML models: logistic regression, decision tree, and random forest, to discriminate between authentic and adulterated honey. To evaluate the performance of the ML models, we use a public dataset comprising measurements of mineral element content of authentic honey, sugar syrups, and adulterated honey. Experimental findings show that mineral element content in honey provides robust discriminative information for detecting honey adulteration. Results also demonstrate that the random forest-based classifier outperforms other classifiers on this dataset, achieving the highest cross-validation accuracy of 98.37%.

## 1 Introduction

Honey is a natural food with high economic value making it a target for adulteration with inexpensive industrial sugar syrups. Honey adulteration aims to increase its volume and gain fast profits. Honey adulteration degrades its quality and has economic and health consequences [1]. Detecting honey adulteration is a challenging problem for consumers since it is difficult to detect adulteration using taste or smell. Several studies have used various technologies for detecting adulteration in honey. Among these technologies are traditional approaches, such as isotopic analysis [3], chromatographic analysis [3], and honey physicochemical parameter analysis [4]. These classical detection methods are time-consuming and require sample preparation and skillful professionals. Besides, the traditional approaches fail to detect adulteration with some industrial sugar syrup. Several researchers used modern methods for detecting honey adulteration. These techniques include various spectroscopic technologies [5-8], hyperspectral imaging [9], electronic nose and electronic tongue [10], and optic fiber sensors [11]. Although the modern techniques have overcome some of the limitations of the traditional methods, they are less accurate and do not provide information about the chemical composition of honey.

A few studies have investigated utilizing minerals with ML models to detect adulteration in honey. Therefore, there is a need to develop robust ML-based techniques for discriminating between authentic and adulterated honey. This study aims to develop an accurate ML model for detecting honey adulteration using mineral element data. In the present paper, we use the Random Forest (RF) algorithm for discriminating between pure and adulterated honey. Besides, we compare the performance of the RF model to the performance of other ML models, such as Logistic Regression (LR) and Decision Tree (DT).

The dataset used in the present study contains measurements of various mineral elements of pure and adulterated honey samples. The botanical origins of the honey samples in the dataset are acacia, chaste, jujube, linden, rape, and Triadica Cochinchinensis (TC). The measured minerals are aluminum, boron, barium, calcium, iron, potassium, magnesium, manganese, sodium, phosphorus, strontium, and zinc. The mineral element content was measured using inductively coupled plasma optical emission spectrometry. The falsified honey samples were obtained by mixing pure honey with sugar syrup at different concentrations. Since some mineral elements were not detected in some sugar syrup and adulterated honey samples, they were labeled as Not Detected (ND), representing missing values in the dataset. The dataset consists of 429 instances and 13 variables, including the class labels. The dataset contains three class labels representing pure honey, sugar syrup, and adulterated honey. Figure 1 depicts the distribution of honey samples in the dataset according to their botanical origins.

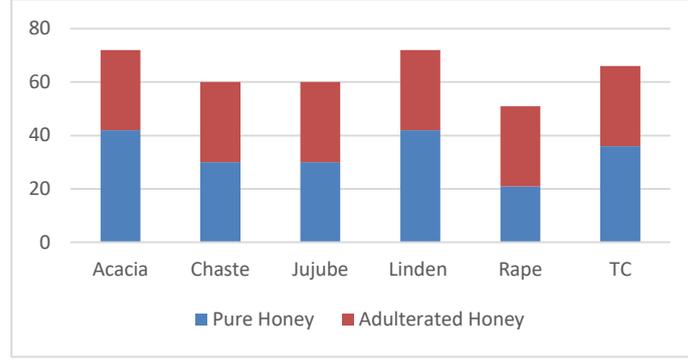

**Fig. 1.** The distribution of honey samples in the dataset according to their botanical origins.

There are two previous works on this dataset. T. Liu et al. [13] used Partial Least Squares-Discriminant Analysis (PLS-DA) to detect adulteration in five monofloral honey types and one multi-floral honey type. The study achieved a classification accuracy of 93% for the mono-floral honey and 87.7% for the multi-floral honey. M. Templ et al. [14] used Linear Discriminant Analysis (LDA), K-Nearest Neighbors (KNN), and Artificial Neural Networks (ANN) to classify the honey samples in the dataset into pure and impure. The study reported that the ANN classifier obtained the lowest misclassification rate.

## 2 Proposed System

The honey adulteration detection method proposed in the present study, depicted in figure 2, comprises two main phases: preprocessing and classification. These two phases are described in detail in the next subsections.

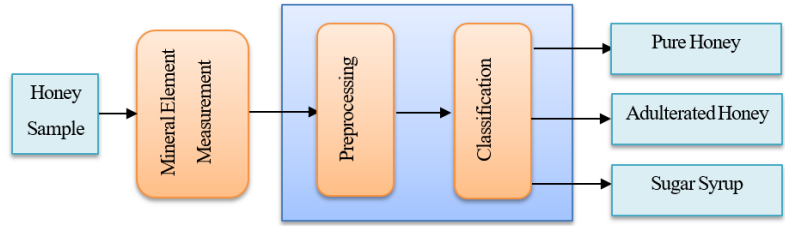

**Fig. 2.** The proposed system's block diagram

### 2.1 Preprocessing

The dataset utilized in the present study contains instances with missing values. The missing values were for the undetected mineral elements in some honey and sugar syrup samples. In this phase, we set the values of the missing-value attributes zeros. We use the min-max normalization approach [15] described by equation 1 to normalize the attributes in the dataset to the interval between zero and one, which helps improve the classification performance of ML models.

$$X_n = \frac{X_o - \min(X_o)}{\max(X_o) - \min(X_o)} \qquad (1)$$

where $X_o$ and $X_n$ are the original and normalized values, respectively.

### 2.2 Classification

We have used three different ML models to discriminate between authentic and adulterated honey in this phase. These models include LR, DT, and RF. We have chosen these three models because they represented linear and nonlinear classifiers and achieved the highest performance compared to other ML classification models. LR is an ML predictive analytic approach that can be used to solve classification tasks based on the probability concept. The independent variables in this model are used to predict the dependent variable and can be measured on a nominal, ordinal, interval, or ratio scale, while the dependent variable can have two or more categories. The dependent variable can have a linear or nonlinear relationship with the independent variables [16]. DT is a supervised machine learning algorithm that can be used to solve both classification and regression problems. Decision tree-based classifiers are easy to interpret, work with numeric and categorical attributes, and can solve binary and multi-class problems [17]. RF is a method for supervised machine learning. It is capable of carrying out classification and regression tasks. The RF algorithm is based on constructing a large number of decision trees. Following their creation, the trees vote for the most popular class [18].

### 2.3 Performance Evaluation

To evaluate the performance of the ML models for discriminating between pure and adulterated honey, we have used the 10-fold cross-validation assessment method. Cross-validation has the advantage of avoiding model overfitting. The performance metrics used were precision, recall, and F1 score. We used these performance metrics since the dataset used in the present research is imbalanced, and the classification accuracy is not sufficient for assessing the performance of the ML models. Precision is the ratio of successfully predicted positive observations to total expected positive observations. The recall is defined as the ratio of correctly predicted positive observations to all observations in the actual class. The F1 score is the weighted average of accuracy and recall. As a result, this score takes both false positives and negatives into account. F1 score is a very useful performance statistic, particularly for imbalanced datasets. Figure 3 illustrates the process of evaluating the performance of ML algorithms.

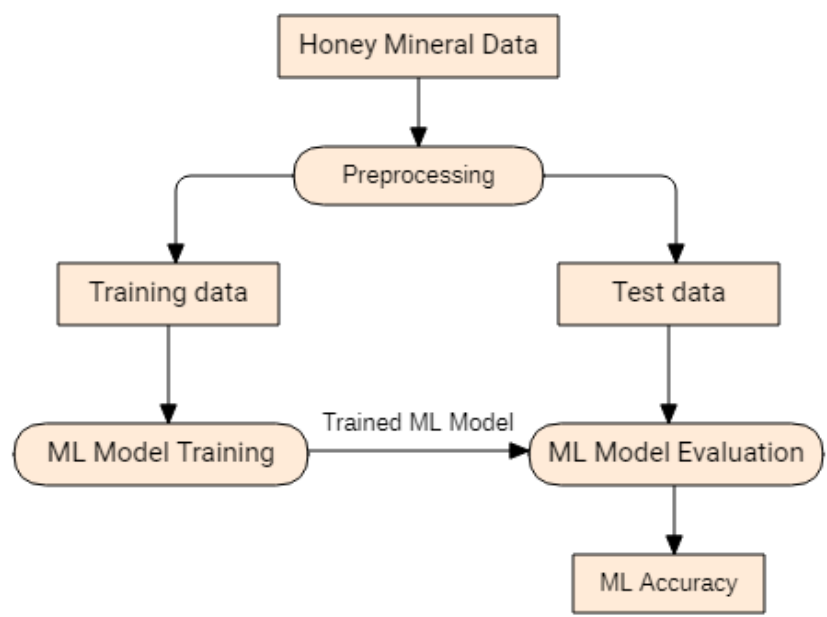

**Fig. 3.** ML models performance evaluation approach

## 3 Results and Discussion

In the present paper, we evaluated the performance of ML models for discriminating between authentic and adulterated honey using two methods. In the first method, we use one ML model for detecting adulteration in different honey types. The second method establishes six ML models for separately detecting adulteration in the six honey types.

### 3.1 General Model for Detecting Honey Adulteration

In this method, we used one ML model for classifying the samples in the dataset into pure honey, adulterated honey, and sugar syrup. We trained the model using all the samples in the dataset. The objective was to assess the effectiveness of ML models in distinguishing authentic honey samples from adulterated samples regardless of the honey botanical origin. In the experiments, we divided the samples in the dataset into three classes. The first and second classes include pure and adulterated honey samples from all floral sources. The third class comprises sugar syrup samples.

**Logistic Regression.** Table 1 shows the various performance metrics of the logistic regression model for the three classes in the dataset. The performance metrics are precision, recall, and F1 score. Table 2 displays the confusion matrix of the model. The confusion matrix gives more information about a predictive model's performance, including which classes are properly predicted, which are wrongly forecasted, and what sort of errors are created. The results show that the logistic regression model perfectly discriminates between honey and sugar syrup. For discriminating between pure and impure honey, the logistic regression model achieved a good recall of 73.1% for pure honey and 68.9% for adulterated honey.

**Table 1.** Performance of Logistic Regression for detecting honey adulteration

| Honey Type | Precision | Recall | F1 Score |
|---|---|---|---|
| Authentic Honey | 0.721 | 0.731 | 0.726 |
| Sugar Syrup | 1.000 | 1.000 | 1.000 |
| Adulterated Honey | 0.700 | 0.689 | 0.694 |

**Table 2.** Confusion matrix of the Logistic Regression classifier

| Classified as | Authentic Honey | Sugar Syrup | Adulterated Honey |
|---|---|---|---|
| Authentic Honey | 147 | 0 | 54 |
| Sugar Syrup | 0 | 45 | 0 |
| Adulterated Honey | 57 | 0 | 126 |

**Decision Tree.** Table 3 displays the performance of the decision tree model for detecting adulteration in honey. The model's confusion matrix is shown in Table 4. The results show an excellent performance of the decision tree for discriminating between authentic honey, sugar syrup, and adulterated honey.

**Table 3.** Performance of Decision Tree for detecting honey adulteration

| Honey Type | Precision | Recall | F1 Score |
|---|---|---|---|
| Authentic Honey | 0.970 | 0.970 | 0.970 |
| Sugar Syrup | 1.000 | 1.000 | 1.000 |
| Adulterated Honey | 0.967 | 0.967 | 0.967 |

**Table 4.** Confusion matrix of the Decision Tree classifier

| Classified as | Authentic Honey | Sugar Syrup | Adulterated Honey |
|---|---|---|---|
| Authentic Honey | 195 | 0 | 6 |
| Sugar Syrup | 0 | 45 | 0 |
| Adulterated Honey | 6 | 0 | 177 |

**Random Forest.** Table 5 displays the performance of the RF model for detecting honey adulteration. Table 6 displays the confusion matrix of the model. The findings show that the RF model obtained excellent performance for detecting adulterated honey, achieving a recall of 98.6%.

**Table 5.** Performance of Random Forest for detecting honey adulteration

| Honey Type | Precision | Recall | F1 Score |
|---|---|---|---|
| Authentic Honey | 0.995 | 0.970 | 0.982 |
| Sugar Syrup | 1.000 | 1.000 | 1.000 |
| Adulterated Honey | 0.968 | 0.986 | 0.981 |

**Table 6.** Confusion matrix of the Random Forest classifier

| Classified as | Authentic Honey | Sugar Syrup | Adulterated Honey |
|---|---|---|---|
| Authentic Honey | 195 | 0 | 6 |
| Sugar Syrup | 0 | 45 | 0 |
| Adulterated Honey | 1 | 0 | 182 |

Table 7 compares the models' performance in discriminating between authentic honey, sugar syrup, and adulterated honey. Results show that the ML models performed well for discriminating between pure and impure honey. Results also show that the RF model outperformed other classifiers, achieving the highest classification accuracy.

**Table 7.** Comparison between ML models performance

| ML Model | Accuracy | Precision | Recall | F1 Score |
|---|---|---|---|---|
| Logistic Regression | 0.741 | 0.741 | 0.741 | 0.741 |
| Decision Tree | 0.972 | 0.972 | 0.972 | 0.972 |
| Random Forest | 0.984 | 0.984 | 0.984 | 0.984 |

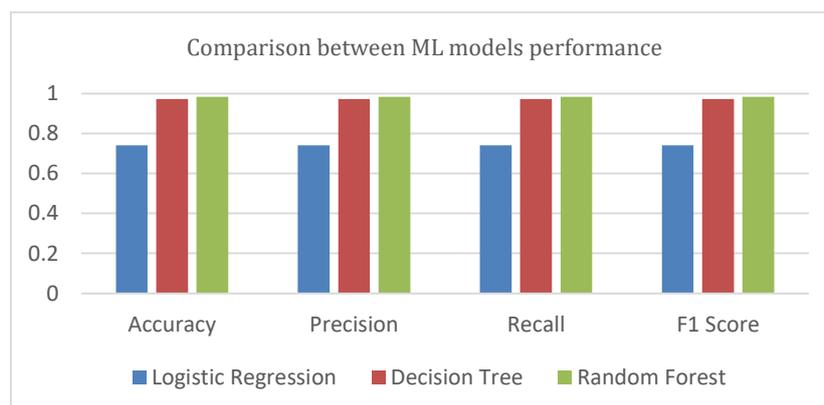

**Fig. 4.** Comparison between ML models' performance for detecting honey adulteration

In this study, we used linear and nonlinear classifiers for discriminating between pure and impure honey. Results show that the Logistic Regression model achieved the lowest detection accuracy, implying that the dataset is linearly inseparable. The Decision Tree achieved accuracy higher than the Logistic Regression since the Decision Tree is a nonlinear classifier. The Random Forest classifier achieved the highest accuracy since it selects the Decision Trees with the highest accuracy. All classifiers successfully distinguished sugar syrup from honey due to the significant, varied mineral element content in honey and the syrup.

### 3.2 Class-wise Adulteration Detection

In this method, we use six ML models for discriminating between pure and impure honey. Each ML model detects adulteration in honey from a specific botanical origin. Table 8 shows the cross-validation accuracy of the ML models for detecting adulteration in the different honey types in the dataset. The results show that all three classifiers achieve good performance, but the RF model achieves the best performance with an average cross-validation accuracy of 99.54%. The RF model perfectly detects adulteration in honey from the chaste, jujube, rape, and TC floral sources. The findings show little improvement in the classifiers' performance in detecting honey adulteration using class-wise detection models.

Like the general detection model, the Logistic Regression obtained the lowest accuracy, and the Random Forest achieved the highest accuracy. The classifiers' detection performance varied according to the honey bo-tanical origin, except the rape honey, where all classifiers perfectly detected the adulteration. The Random Forest classifier perfectly detected the impurity in the jujube honey, rape honey, and TC honey.

**Table 8.** The cross-validation accuracy of the ML models for detecting adulteration in different honey types

| Botanical Origin | Acacia | Chaste | Jujube | Linden | Rape | TC | Average |
|---|---|---|---|---|---|---|---|
| Logistic Regression | 91.67 | 88.33 | 96.67 | 87.5 | 100 | 92.42 | 92.77 |
| Decision Tree | 97.22 | 91.67 | 88.33 | 95.83 | 100 | 98.48 | 95.26 |
| Random Forest | 98.61 | 100 | 100 | 98.61 | 100 | 100 | 99.54 |

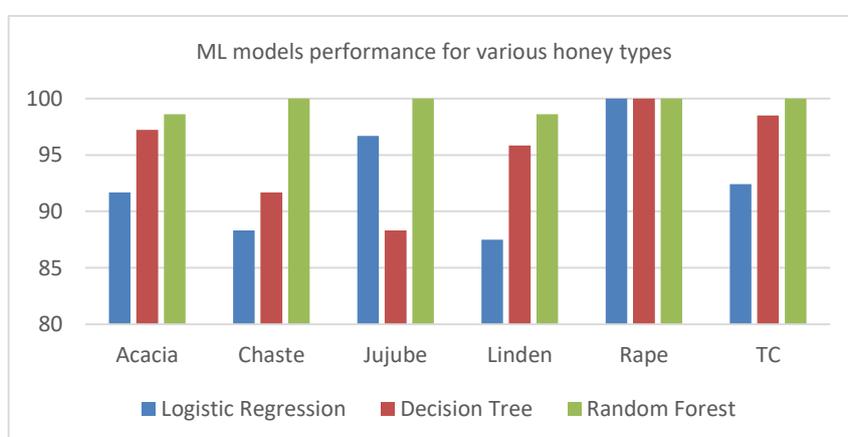

**Fig. 5.** ML models performance for various honey types

### 3.3 Mineral Element Significance for Detecting Honey Adulteration

Figure 6 shows the significance of the mineral elements for detecting honey adulteration according to the Random Forest classifier. The figure reveals that barium, boron, sink, potassium, and iron are the most significant mineral elements for discriminating between authentic and adulterated honey. The barium was the most significant attribute since this mineral was absent in most authentic honey samples.

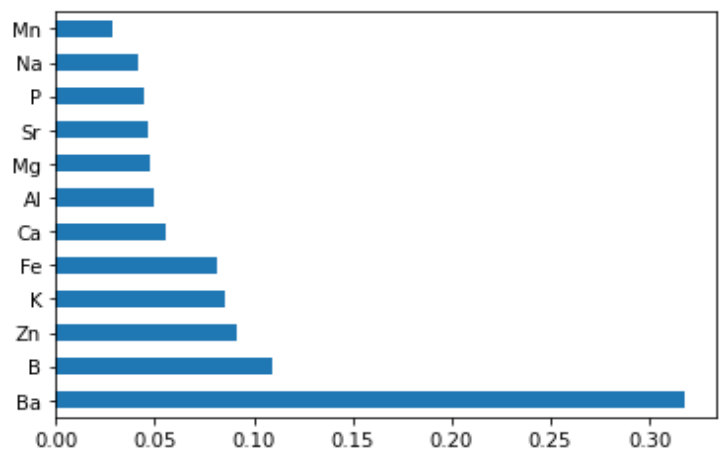

**Fig. 6.** Mineral element significance for detecting honey adulteration

# 4     Conclusions

Distinguishing genuine honey from adulterated honey is a challenging problem for consumers. This paper proposed an ML-based system for detecting adulterated honey using mineral element profiles. Experimental findings show that mineral element content provides discriminative information helpful for detecting honey adulteration. Experimental results also demonstrate that the RF classifier achieved the best performance, outperforming other ML classifiers. The performance of the ML models for detecting honey adulteration varied according to the honey's botanical origin. Among six honey botanical origins, the adulteration was accurately detected in the chaste, jujube, rape, and TC honey. Moreover, this study concludes that ML models in combination with mineral element profiles can effectively discriminate between pure and impure honey regardless of the honey's botanical origin.

**Acknowledgments.** This work was supported by the Department of Science and Technology under the Funds for Infrastructure through Science and Technology (DST-FIST) grant SR/FST/ETI-340/2013 to the Department of Computer Science and Information Technology at Dr. Babasaheb Ambedkar Marathwada University in Aurangabad, Maharashtra, India. The authors would like to express their gratitude to the department and university administrators for providing the research facilities and assistance.